\documentclass{article}
\usepackage{ijcai25}

\usepackage{times}
\usepackage{soul}
\usepackage{url}
\usepackage[hidelinks]{hyperref}
\usepackage[utf8]{inputenc}
\usepackage[small]{caption}
\usepackage{graphicx}
\usepackage{amsmath}
\usepackage{amsthm}
\usepackage{booktabs}
\usepackage{algorithm}
\usepackage{algorithmic}
\usepackage[switch]{lineno}

\usepackage{pifont}
\usepackage{caption}
\usepackage{multirow}
\usepackage{amssymb}
\usepackage{arydshln}
\usepackage{fix-cm}

\title{FedAPA: Server-side Gradient-Based Adaptive Personalized Aggregation for Federated Learning on Heterogeneous Data}

\author{}

\author{
Yuxia Sun$^1$
\and
Aoxiang Sun$^2$\and
Siyi Pan$^{1}$\and
Zhixiao Fu$^3$\and
Jingcai Guo*$^4$\\
\affiliations
$^1$College of Information Science and Technology, Jinan University\\
$^2$College of Information and Computational Science, Jilin University\\
$^3$Department of Computer Science, University of
Toronto\\
$^4$Department of Computing, The Hong Kong Polytechnic University\\
\emails
tyxsun@email.jnu.edu.cn,
2835604014@qq.com,
StuPsy@stu2022.jnu.edu.cn,\\
zx.fu@mail.utoronto.ca,
jc-jingcai.guo@polyu.edu.hk (\footnote{Jingcai Guo is the corresponding author}corresponding author)
}

\begin{document}

\maketitle

\begin{abstract}
Personalized federated learning (PFL) tailors models to clients' unique data distributions while preserving privacy. However, existing aggregation-weight-based PFL methods often struggle with heterogeneous data, facing challenges in accuracy, computational efficiency, and communication overhead. We propose FedAPA, a novel PFL method featuring a server-side, gradient-based adaptive aggregation strategy to generate personalized models, by updating aggregation weights based on gradients of client-parameter changes with respect to the aggregation weights in a centralized manner. FedAPA guarantees theoretical convergence and achieves superior accuracy and computational efficiency compared to 10 PFL competitors across three datasets, with competitive communication overhead.
\end{abstract}

\section{Introduction}

\begin{figure}[t]
    \centering
    \includegraphics[width=0.9\linewidth]{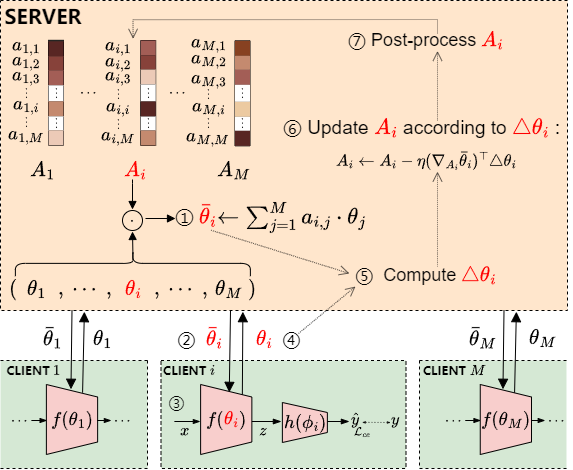}
    \caption{Framework of FedAPA. \ding{172} The server generates personalized model $\bar{\theta}_i$ via aggregation according to weight vector $A_i$; \ding{173} Each client downloads its model $\bar{\theta}_i$; \ding{174} Local training on private data; \ding{175} Each client upload its model $\theta_i$; \ding{176} The server computes the update of model parameters $\triangle\theta_i$; \ding{177} Updates $A_i$ via gradient descent according to $\triangle\theta_i$; \ding{178} Post-processes $A_i$ via clipping, self-weight setting, and normalization.}
    \captionsetup{justification=centering}
    \label{Fig1 Framework.png}
\end{figure}

As a prominent distributed machine learning paradigm, Federated Learning (FL) allows clients to collaboratively train models while keeping data local \cite{10.1145/3298981}. In cross-silo FL, client data is often non-IID (non-Independent and Identically Distributed), exhibiting statistical heterogeneity that leads to client drift and reduces the effectiveness of methods like FedAvg \cite{pmlr-v54-mcmahan17a}.

To address the challenges posed by heterogeneous data distributions, Personalized Federated Learning (PFL) has been developed to tailor models to the unique data characteristics of individual clients\cite{9090366}. PFL can be broadly categorized into three main approaches: (1) Personalized model-based approaches, which directly create a personalized local model for each client, typically without relying on a global model \cite{10.1007/978-3-030-86486-6_36,Huang_Chu_Zhou_Wang_Liu_Pei_Zhang_2021,ijcai2022p301,pmlr-v139-shamsian21a}; (2) Global model-based approaches, where a global model is first constructed and then personalized through techniques like client-specific fine-tuning \cite{xu2023personalized,NEURIPS2023_2e0d3c6a}; and (3) Other PFL approaches include those that leverage information beyond model parameters, such as prototypes,
to optimize local models \cite{Tan_Long_LIU_Zhou_Lu_Jiang_Zhang_2022}. Among these, personalized model-based approaches, the primary focus of this paper, have gained significant attention for their ability to directly tailor local models to each client’s unique data distribution.


We categorize existing personalized model-based PFL approaches into three groups based on their aggregation strategies for creating each personalized model: 
(1) Static metric-based methods: Approaches like FedAMP\cite{Huang_Chu_Zhou_Wang_Liu_Pei_Zhang_2021} and FedPHP\cite{10.1007/978-3-030-86486-6_36} define aggregation weights using predefined metrics derived from parameter similarity or training progress. However, these predefined metrics do not directly reflect client-specific optimization objectives (e.g., local losses or gradients).
(2) Local gradient-based methods: Approaches such as APPLE\cite{ijcai2022p301} and FedALA\cite{Zhang_Hua_Wang_Song_Xue_Ma_Guan_2023} calculate aggregation weights locally via gradient descent. While these methods share model parameters between clients or use element-wise model combination, they incur high computational complexity and potential communication overhead. Additionally, their focus on individual client objectives limits effective cross-client coordination.
(3) Auxiliary network-based methods: Approaches like pFedHN\cite{pmlr-v139-shamsian21a} 
utilize Hypernetworks\cite{DBLP:journals/corr/HaDL16} to derive personalized models or aggregation weights. However, the need to train auxiliary models introduces significant computational overhead. In summary, these limitations highlight the need for new model-aggregation techniques that not only enable adaptive personalization in non-IID scenarios but also reduce computational and communication costs.

To address these limitations, we propose a novel personaliezed model based PFL approach, \textbf{FedAPA} (Adaptive Personalized Aggregation for Federated Learning), as illustrated in Fig.~\ref{Fig1 Framework.png}. FedAPA introduces a server-side aggregation strategy that adaptively learns aggregation weights using gradients of client parameter changes with respect to the aggregation weights, generating aggregated personalized models for each client. Specifically, the server: (1) maintains the client parameters and an \textbf{aggregation-weight vector}, where each element quantifies the relevance of other clients' model knowledge during aggregation; (2) constructs personalized client parameters through weighted aggregation of all collaborative clients' parameters using the learned weights; and (3) updates the weight vector based on the \textbf{client-parameter delta}—the change in the client’s local model parameters after multiple local update steps. As an additional feature, FedAPA adopts a partial model sharing strategy, transferring only the client’s feature extractor to reduce irrelevant information during aggregation.


The key innovation of FedAPA lies in its aggregation strategy, which updates aggregation weights on the server side using the gradient of the client-parameter delta with respect to the aggregation weights, optimizing the personalized aggregation process. Half of the squared norm of the client-parameter delta serves as a proxy for the local loss, effectively capturing each client’s optimization state. By dynamically adjusting aggregation weights to reduce this proxy loss, FedAPA generates an aggregated personalized model for each client, better aligning the model with the client’s local objectives. Compared to existing model-aggregation approaches, FedAPA’s strategy offers the following advantages: (1) Adaptability: FedAPA dynamically learns aggregation weights based on gradients derived from the proxy loss, enabling the aggregation weights to adapt effectively to each client’s unique optimization trajectory. 
(2)  Centralized Efficiency: By centralizing weight updates on the server, FedAPA minimizes communication and computational overhead while enhancing cross-client coordination. (3) Simplicity: FedAPA avoids the need for auxiliary networks, eliminating the computational complexity associated with training additional models. In summary, FedAPA’s aggregation strategy effectively balances personalization and collaboration in non-IID scenarios, enhancing computational and communication efficiency.

We summarize our contributions as follows:
\begin{itemize}
    \item We propose FedAPA, a novel model-aggregation PFL framework for heterogeneous data scenarios, featuring a server-side gradient-based model-aggregation strategy that enhances FL performance while maintaining low computational and communication overhead.
    while maintaining low computational and communication overhead.
    \item To the best of our knowledge, FedAPA is the first to adaptively update aggregation weights on the server side to generate aggregated personalized models, using gradients derived from client-parameter changes.
    \item We provide a theoretical convergence guarantee for the FedAPA algorithm.
    \item We experimentally demonstrate that our approach outperforms 10 PFL competitors across three benchmarks under two non-IID settings, achieving top accuracy, optimal computational efficiency, and low communication overhead. Ablation studies highlight the critical contribution of the aggregation strategy and the complementary benefits of the partial model sharing strategy.
\end{itemize}

\section{Related Work}

\subsection{Personalized model-based PFL Approaches}
Personalized model-based PFL approaches, also known as PFL approaches via learning personalized local models, directly create personalized models for individual clients, typically without the need for a global model. These approaches employ various aggregation strategies to construct each personalized model, which can be broadly categorized as follows:
(1)	Static metric-based methods: These methods define aggregation weights using predefined, static metrics. For example, FedAMP \cite{Huang_Chu_Zhou_Wang_Liu_Pei_Zhang_2021} calculates similarity as weights based on the Euclidean distance of model parameters, while FedPHP \cite{10.1007/978-3-030-86486-6_36} measures relevance using training progress. 
Although straightforward, these metrics do not directly align with client-specific optimization objectives, limiting their adaptability to non-IID data distributions.
(2) Local gradient-based methods: They learn aggregation weights via gradient descent on the client side. For example, APPLE \cite{ijcai2022p301} learns inter-client relations by sharing model parameters between clients, leading to high communication costs and privacy concerns. FedALA \cite{Zhang_Hua_Wang_Song_Xue_Ma_Guan_2023} computes element-wise combination weights between client and global models, resulting in high computational complexity. While these methods utilize non-local information, their focus on local objectives limits cross-client coordination and adaptability to data heterogeneity.
(3) Hypernetwork-based methods: Hypernetworks \cite{DBLP:journals/corr/HaDL16} have been utilized as the auxilliary networks to enhance personalization. For example, pFedHN \cite{pmlr-v139-shamsian21a} employs Hypernetworks to generate personalized models
However, these methods incur significant computational overhead due to the complexity of training Hypernetworks

\subsection{ Global model-based PFL Approaches}
Global model-based PFL approaches, also referred to as PFL approaches via global model personalization, involve training a global model first and then personalizing it to derive local models for individual clients. For example, 
FedPAC \cite{xu2023personalized} leverages a probabilistic framework for personalization using Bayesian neural networks and a global model, allowing personalized models to adaptively incorporate uncertainty during client-specific training. DBE \cite{NEURIPS2023_2e0d3c6a} enhances the generalization of the global feature extractor through bi-directional knowledge transfer, reducing representation bias and fine-tuning client models for improved personalization.

\subsection{Other PFL Approaches}
PFL can also leverage non-model information, such as prototypes, to achieve personalization. For example, FedProto \cite{Tan_Long_LIU_Zhou_Lu_Jiang_Zhang_2022} leverages local prototypes to construct global prototypes, which are used to improve feature extraction and classification. As a combined method, FedGH\cite{10.1145/3581783.3611781} uses client-uploaded prototypes (i.e., class-averaged representations) to train a shared global head on the server, which is then downloaded to replace local heads.

\section{Method}
\subsection{Problem Formulation}

Personalized Federated Learning (PFL) aims to collaboratively train personalized models for a set of clients, each with its own private data. PFL addresses the challenge of data heterogeneity across clients, ensuring that each client's model is tailored to its specific data while maintaining data privacy and improving individual model performance.

Let $M$ be the total number of clients participating in the federated learning process. Each client $i \in \{1,\cdots,M\}$ has a local dataset $D_i$; with its own data distribution $P_i$ on $X \times Y$. In PFL, each client $i$ seeks to learn a personalized model parameterized by $\omega_i$. The objective function for PFL can be formulated as:
\begin{equation}
    \arg\underset{\omega_1,\cdots,\omega_M}{\min} \sum_{i=1}^{M} \frac{\left|D_i\right|}{N} L_i \left(F\left(\omega_i;x\right), y\right)
\end{equation}
\noindent where $\left|D_i\right|$ is the size of the local dataset $Di$; $N$ is the total number of data points across all clients; $F\left(\omega_i;x\right)$ is the model parameterized by $\omega_i$ evaluated on input $x$; $y$ is the true label for input $x$; and $L_i$ is the local loss function for client $i$, such as the following cross-entropy loss function:
\begin{multline}
    L_{CE,i}\left(F\left(\omega_{i};x\right),y\right)\\
    = -\frac{1}{\left|D_i\right|}\sum_{\left(x,y\right) \in D_i}\sum_{c=1}^{C}y_c\log\left(F\left(\omega_i;x\right)_c\right)
\end{multline}
\noindent where $C$ is the number of classes; $y_c$ is the binary indicator (0 or 1) if class label $c$ is the correct classification for input $x$; and $F\left(\omega_i;x\right)_c$ is the predicted probability of input $x$ belonging to class $c$ under the model parameterized by $\omega_i$.

\subsection{FedAPA Algorithm}
The framework of FedAPA is demonstrated in Fig\ref{Fig1 Framework.png}. Each FedAPA client uploads its feature extractor, receives a personalized feature extractor after server-side aggregation, and then updates the entire client model locally, including both the feature extractor and the classifier. On the server side, the model parameters of each client's feature extractor and the corresponding aggregation-weight vector are maintained. The core of FedAPA lies in the server-side aggregation, where the server constructs a personalized feature extractor for each client by adaptively aggregating all clients' model parameters using weights updated via gradient descent.

FedAPA adjusts each aggregation weight vector on the server using the gradient calculated from the client parameter delta, which represents the change in local model parameters after multiple local updates within a communication round. At the end of each round, the updated client parameters can be considered as an approximation of the local task's optimal solution, with the client-parameter delta approximately reflecting the local optimization direction (as detailed in subsubsection \ref{subsec:3.2.4}). For each client, the server iteratively updates the client's aggregation-weight vector based on the gradient of the client-parameter delta with respect to the vector. The learned weights represent adaptively acquired knowledge contributions from other clients, which may potentially reflect the data distribution correlation among clients. By aggregating beneficial model knowledge from collaborative clients, FedAPA addresses non-IID data challenges for individual clients effectively.


\subsubsection{Local Training}
As shown in Fig\ref{Fig1 Framework.png}, for each client $i$, the local model $F(\omega_i)$ is partitioned into two parts: the feature extractor $f(\theta_i)$, which takes the input sample $x$ and produces the representation $z = f(\theta_i;x)$, and  the decision layer $h(\phi_i)$, which processes the representation $z$ and outputs the predicted label $\hat{y} = h(\phi_i;z)$. The shared parameter $\theta_i$ is uploaded to the server, while the private parameter $\phi_i$ is kept locally. Thus, the local objective function in PFL can be expressed as 
\begin{equation}
\begin{aligned}
    \arg \underset{\omega_i}{\min} \mathcal{L}_i\left(F\left(\omega_i;x\right), y\right) =  \arg \underset{\theta_i,\phi_i}{\min} \mathcal{L}_i(h(\phi_i;f(\theta_i;x)),y)
\end{aligned}
\end{equation}
Upon downloading the parameters $\bar{\theta}_i$ as the current feature extractor, client $i$ obtains its current entire model $\omega_i := [\theta_i;\phi_i]$. Each client conducts several local update steps on the local data using the following update rule:
\begin{equation}
\begin{aligned}
    \omega_i^{(t+1)} \leftarrow \omega_i^{(t)} - \alpha_i \nabla_{\omega_i}L_i\left(F(\omega_i^{(t)};x), y\right)
\end{aligned}
\end{equation}
where $\alpha_i$ denotes the learning rate of the local model parameters $\omega_i$. After the local updates, client $i$ separates $\theta_i$ from the updated $\omega_i$, and uploads the updated $\theta_i$ as the shared client parameters to the server.


\subsubsection{Server-Side Aggregation of Client-Parameters}
For each client $i$, FedAPA server manages a trainable aggregation-weight vector $A_i = (a_{i,1}, \dots, a_{i,j}, \dots, a_{i,M})^T$ where $i,j \in \{1,\cdots,M\}$, 
and $M$ indicates the total number of clients in the federated learning system. Here, $a_{i,j}$ denotes the weight of client $j$ during the weighted aggregation used to form the target parameters $\bar{\theta}_i$ for client $i$. During the initialization of the learning process, we set the self-weight $a_{i,i}$ to 1, while the other weights $a_{i,j}$ to 0 for $i \ne j$. For the $t$-th training round, the server computes the personalized parameters $\bar{\theta}_i$ for client $i$ by performing a weighted aggregation of all clients’ shared parameters according to the following formula:
\begin{equation}
\begin{aligned}
    \bar{\theta}_{i} = \sum_{j = 1}^M a_{i, j} \cdot \theta_j
    \label{con:2}
\end{aligned}
\end{equation}

\subsubsection{Server-Side Updating of Weight-Vectors} \label{subsec:3.2.4}
Upon receiving the client-parameters $\theta_i$  from each client $i$, the FedAPA server calculates the difference between the current $\theta_i$ and the previous parameter versions $\bar{\theta}_i$ stored on the server, obtaining the client-parameter delta as follows:
\begin{equation}
\begin{aligned}
    \Delta \theta_i = \theta_{i} - \bar{\theta}_{i}
\end{aligned}
\end{equation}
Subsequently, the server leverages this client-parameter delta to update the aggregation-weight vector $A_i$ for client $i$ according to the following rule:
\begin{equation}
\begin{aligned}
    A_i \leftarrow A_i - \eta(\nabla_{A_i}\bar{\theta}_{i})^T\Delta\theta_i
    \label{Formula (7)}
\end{aligned}
\end{equation}
where $\eta$ is the learning rate for updating the aggregation-weight vector on the server.
\begin{algorithm}[tb]
    \caption{FedAPA}
    \label{alg:pseudocode}
    \textbf{Input}: dataset $\{\mathcal{D}_1,  \dots, \mathcal{D}_M\}$, learning rates $\eta$ and $\alpha_i$, total communication rounds $T$,  local rounds $E$, client-parameter set $\{\theta_1,  \dots, \theta_M\}$, aggregation-weight vector set $\{A_1, \dots, A_M\}$.\\
    \textbf{Output}: Trained personalized model set $\{{\bar{\theta}_{1}}, \bar{\theta}_{2}, \dots, \bar{\theta}_{M}\}$.
    \begin{algorithmic}[1]
        \STATE $\forall i\in\{1,\cdots,M\}$, initialize $\theta_i$ on server
        \STATE $\forall i\in\{1,\cdots,M\}$, initialize $A_i = (a_{i,1},  \dots, a_{i,M})^T$  on server
        
        \STATE \textbf{Function} Server Executes
            \FOR{each communication round $t = 1, \dots, T$} 
                \STATE Sample a subset of clients $S^t$
                \FOR{each client $i \in S^t$ \textbf{in parallel}} 
                    \STATE $\bar{\theta}_{i} \gets \sum_{j = 1}^M a_{i, j}\theta_{j}$ \COMMENT{\textbf{Weighted aggregation}}
                    \STATE $\theta_{i} \gets  ClientUpdate(\bar{\theta}_{i})$
                    \STATE $\Delta \theta_i \gets \theta_{i} - \bar{\theta}_{i}$ 
                    \COMMENT{\textbf{Computing $\Delta\theta_i$}}
                    \STATE Update $A_i$ via Eq \ref{Formula (7)} 
                    \COMMENT{\textbf{Updating based on $\Delta \theta_i$}}
                    \STATE $a_{i,j} \gets min(max(a_{i,j}, 0), 1)$
                    , $j = 1, \dots, M$
                    \COMMENT{\textbf{Clipping}}
                    \STATE $a_{i,i} \gets 0.5$
                    \STATE Normalize $A_i$ 
                \ENDFOR
            \ENDFOR
        \STATE \textbf{End Function}
        \STATE \textbf{Function} Clientupdate($\bar{\theta}_{i}$) 
            \STATE Download $\bar{\theta}_{i}$
            \STATE $\theta_{i} \gets \bar{\theta}_{i}$.
            \STATE $\omega_i \gets [\theta_i;\phi_i]$
                \FOR{each local epoch $e = {1, \dots, E}$} 
                    \FOR{mini-batch $B \subseteq \mathcal{D}_i$} 
                        \STATE $\omega_i \gets \omega_i - \alpha_i\nabla_{\omega_i}L_i(B)$
                        \COMMENT{\textbf{Local training}}
                    \ENDFOR
                \ENDFOR
            \STATE $[\theta_i;\phi_i] \gets \omega_i$
            \STATE \textbf{return} $\theta_i$ \COMMENT{\textbf{Uploading $\theta_i$}}
        \STATE \textbf{End Function}
    \end{algorithmic}
\end{algorithm}

\textbf{Equation (7) is pivotal to the FedAPA algorithm.} To provide intuition behind this formula, consider the optimal solution for the local task's model parameters, $\theta^*_i = \arg \underset{\theta_i} \min L_i$. After multiple local updates within a communication round, $\theta_i$ can be regarded as an approximation of $\theta^*_i$. Consequently, $\frac{1}{2} \|\theta^*_i - \bar{\theta}_i\|_2^2$ approximates $\frac{1}{2} \|\theta_i - \bar{\theta}_i\|_2^2$, which is equivalent to half of the squared norm of $\Delta \theta_i$. This measure, $\frac{1}{2} \|\theta^*_i - \bar{\theta}_i\|_2^2 = \frac{1}{2} \|\bar{\theta}_i - \theta^*_i\|_2^2$, can be interpreted as a surrogate for the local loss. Thus, half of the squared norm of $\Delta \theta_i$ serves as a proxy for the local loss $L_i$, where $\Delta \theta_i$ can be interpreted as $-\nabla_{\bar{\theta}_i} L_i$. According to the chain rule, \((\nabla_{A_i}\bar{\theta}_{i})^T\Delta\theta_i\) represents \(-\nabla_{A_i} L_i\). Consequently, Equation (7) updates the aggregation weights $A_i$ on the server in a way that minimizes the local loss $L_i$ for each client, ensuring that the personalized aggregation aligns with the client-specific optimization trajectory.

Before sending the updated aggregation-weight vector $A_i$ to client $i$ , the server further optimizes the weights using following three post-processing steps:
(1) Clipping: To improve training stability and convergence, we control the update range of each weight element in the vector $A_i$ by clipping each element value $a_{i,j}$ to be within $[0, 1]$, where $j \in \{1,\cdots,M\}$. (2) Self-weight adjusting: Each client should prioritize its own local knowledge over that of other clients. Therefore, for the clipped aggregation-weight vector of each client $i$, we set the weight of the element $a_{i,i}$ (also known as the self-weight $\mu$) to 0.5, ensuring it is not less than the weights of the other elements in the vector. (3) Normalization: To balance client contributions and minimize variance during aggregation, thus improving training generalization, we normalize each aggregation-weight vector from the previous step to ensure that the total sum of the weights equals 1.

In summary, during a communication round for each client $i$, the FedAPA server first constructs the personalized client parameters $\bar{\theta}_{i}$ by performing a weighted aggregation of the collaborative clients’ parameters. Next, the client downloads $\bar{\theta}_{i}$ and uploads the optimized client parameters $\theta_i$ after local training. Subsequently, the server computes the client-parameters change $\Delta\theta_i$, and uses it to update the aggregation-weight vector $A_i$, adaptively learning beneficial knowledge from collaborative clients. Algorithm 1 outlines the overall learning process in FedAPA.

\section{Convergence Analysis}
In this section, we analyze the convergence of FedAPA under suitable conditions. Additional notations used in this section are detailed in Appendix.

\noindent\textbf{Assumption 1. }\hypertarget{A:1}
($Lipschitz$ Smooth). Each local objective function is $L_1$-$Lipschitz$ smooth, which also means the gradient of local objective function $\mathcal{L}$ is $L_1$-$Lipschitz$ continuous. This assumption is reasonable as it encapsulates the differentiability and smoothness properties inherent to many common loss functions in deep learning, aligning with the mathematical prerequisites for the convergence of gradient-based optimization methods fundamental to neural network training,
\begin{equation}
\begin{aligned}
    ||\nabla \mathcal{L}_{t_1} - \nabla \mathcal{L}_{t_2}|| 
    &\le L_1||\omega_{i}^{(t_1)} - \omega_{i}^{(t_2)}||_2, \\
    &\forall t_1,t_2 > 0, i \in \{1,2,\dots,M\}.
\end{aligned}
\end{equation}
which implies the following quadratic bound,
\begin{align}
    &\mathcal{L}_{t_1} - \mathcal{L}_{t_2} \le \langle \nabla \mathcal{L}_{t_2}, (\omega_{i}^{(t_1)} - \omega_{i}^{(t_2)}) \rangle\\
    & + \frac{L_1}{2}||\omega_{i}^{(t_1)} - \omega_{i}^{(t_2)}||_2^2, \nonumber \forall t_1,t_2 > 0, i \in \{1,2,\dots,M\}.
\end{align}

\noindent\textbf{Assumption 2. }\hypertarget{A:2}
(Unbiased Gradient and Bounded Variance). The stochastic gradient $gr_{i}^{(t)}$ = $\nabla \mathcal{L}(\omega_{i}^{(t)}, \xi_i)$ is an unbiased estimator of the local gradient for each client, where the random variable $\xi_i$ follows the distribution $\mathcal{D}_i$. This foundational assumption is pivotal, as it ensures that the expectation of the stochastic gradient, when averaged over the local dataset, corresponds to the true gradient of the local loss function. Despite the inherent variability in data distribution across different clients, this property remains intact, underscoring its relevance in Federated Learning where data heterogeneity is often the rule rather than the exception,
\begin{equation}
\begin{aligned}
    \mathbb{E}_{\xi_i \sim \mathcal{D}_i}[gr_{i}^{(t)}] = \nabla \mathcal{L}(\omega_{i}^{(t)}) = \nabla \mathcal{L}_t, \forall i \in \{1,2,\dots,M\},
\end{aligned}
\end{equation}
and its variance is bounded by the constant $\sigma^2$:
\begin{equation}
\begin{aligned}
    \mathbb{E}[||gr_{i}^{(t)} - \nabla \mathcal{L}(\omega_{i}^{(t)})||_2^2] &\le \sigma^2, \\
    &\forall i \in \{1,2,\dots,M\}, \sigma^2\ge 0.
\end{aligned}
\end{equation}

\noindent\textbf{Assumption 3. }\hypertarget{A:3}
($Lipschitz$ Continuity). Each local loss function $\mathcal{L}$ is $L_2$-$Lipschitz$ continuous with respect to $\theta$. This assumption is deemed reasonable, akin to assumption 1, as it reflects a common property of loss functions, ensuring the boundedness of the gradient's impact on parameter updates, which is a prerequisite for the stability and convergence of optimization algorithms,
\begin{equation}
\begin{aligned}
    ||\mathcal{L}(\theta_{i}^{(t_1)}) - \mathcal{L}(\theta_{i}^{(t_2)})|| \le &L_2||\theta_{i}^{(t_1)} - \theta_{i}^{(t_2)}||_2,\\
    &\forall t_1,t_2 > 0, i \in \{1,2,\dots,M\}.
\end{aligned}
\end{equation}

\noindent \textbf{Theorem 1.} (One-round Deviation Bound).\hypertarget{T:1}{}
Let Assumptions 1 to 3 hold. For an arbitrary client, after every communication round, we have,
\begin{equation}
\begin{aligned}
        \mathbb{E}[\mathcal{L}_{(t + 1)E + \frac{1}{2}}] 
        &\le \mathbb{E}[\mathcal{L}_{tE + \frac{1}{2}}] - (\alpha - \frac{L_1 \alpha ^2}{2}) \sum_{e = \frac{1}{2}}^{E - 1} || \nabla \mathcal{L}_{tE + e} ||_2^2 \\
        &+ \frac{L_1 E \alpha ^2}{2} \sigma^2 + 2L_2\eta (t + 1)Max^3
\end{aligned}
\end{equation}  
where $Max = \underset{t = 1, 2, \dots}{\underset{i = 1, 2, \dots, M}{max}} ||\theta_i^{(t)}||$.

This theorem indicates the deviation bound of the local objective function for an arbitrary client after each communication round. Convergence of each client's local objective function can be guaranteed when there is a certain expected one-round decrease, which can be achieved by choosing appropriate $\eta$ and $\alpha$. \textbf{The proof of Theorem 1 is in Appendix.}

\begin{table*}[t]
\begin{center}

\fontsize{9pt}{8.5pt}\selectfont
\begin{tabular}{ c | l | c  c  c | c  c  c }
\hline
 & & \\[-7pt]
\multicolumn{2}{c|}{\multirow{4}{*}{Approach}} & \multicolumn{3}{c|}{\textbf{Pathological Non-IID}} & \multicolumn{3}{c}{\textbf{Practical Non-IID}}\\
\cline{3-8}
 & & & & & & \\[-7pt]
\multicolumn{2}{c|}{} & FMNIST & CIFAR-10 & CIFAR-100 & FMNIST & CIFAR-10 & CIFAR-100\\
\multicolumn{2}{c|}{} & (Simple) & (Medium) & (Complex) & (Simple) & (Medium) & (Complex)\\
\cline{3-8}
 & & & & & & \\[-7pt]
\multicolumn{2}{c|}{} & $c = 2$ & $c = 2$ & $c = 10$ & $\alpha = 0.1$ & $\alpha = 0.1$ & $\alpha = 0.01$\\
\hline

\multirow{11}{*}{\rotatebox{90}{\#Client = 20}} & FedAvg~\cite{pmlr-v54-mcmahan17a} & 75.08 & 48.48 & 18.49 & 86.96 & 56.68 & 20.39\\
 & FedPHP~\cite{10.1007/978-3-030-86486-6_36} & 99.17 & 84.40 & 52.02 & 95.33 & 82.52 & 62.09\\
 & FedAMP~\cite{Huang_Chu_Zhou_Wang_Liu_Pei_Zhang_2021} & 99.26 & 86.31 & 56.94 & {96.44} & 82.38 & 66.43\\
 & APPLE~\cite{ijcai2022p301} & 99.06 & 84.69 & 54.67 & 95.42 & 79.79 & 61.83\\
 & FedALA~\cite{Zhang_Hua_Wang_Song_Xue_Ma_Guan_2023} & {99.33}  & {87.52} & \textbf{59.75} & 96.19 & 82.53 & {67.24}\\
& pFedHN~\cite{pmlr-v139-shamsian21a} & 99.29 & 86.12 & 57.43 & 95.95 & 80.71 & 64.55\\
& pFedLA~\cite{Ma_2022_CVPR} &98.85 &78.85 &41.96 &95.30 &82.86 &50.02 \\[1pt]
\cdashline{2-8} 
 & & & & & & & \\[-7pt]
& FedPAC~\cite{xu2023personalized}&\textbf{99.41}&86.18&56.06&96.36&84.93&61.89\\
    & DBE~\cite{NEURIPS2023_2e0d3c6a} & 99.20 & 86.39 & 57.46 & 95.97 & {83.45} & 
  63.00\\
    & FedProto~\cite{Tan_Long_LIU_Zhou_Lu_Jiang_Zhang_2022} & 99.31 & 86.39 & 57.20 & 95.52 & 82.71 & 
  66.81\\
  & FedGH~\cite{10.1145/3581783.3611781} & 99.32 & 86.50 & 57.34 & 95.22 & 80.75 & 
  66.59\\ 
 \cline{2-8} 
 & & & & & & & \\[-7pt]
 & \text{\textbf{FedAPA \textit{(ours)}}} & \textbf{99.35} & \textbf{87.69} & \textbf{59.24} & \textbf{96.71} & \textbf{85.85} & \textbf{68.11}\\ 
\hline 
 & & & & & & & \\[-8pt]
\multirow{11}{*}{\rotatebox{90}{\#Client = 50}} & FedAvg~\cite{pmlr-v54-mcmahan17a} & 72.54 & 38.69 & 13.19 & 82.98 & 50.77 & 11.84\\
 & FedPHP~\cite{10.1007/978-3-030-86486-6_36} & 98.90 & 82.85 & 46.20 & 95.27 & 81.73 & 69.33\\
 & FedAMP~\cite{Huang_Chu_Zhou_Wang_Liu_Pei_Zhang_2021} & 99.02 & 83.56 & 49.61 & 94.40 & 82.18 & 73.67\\
  & APPLE~\cite{ijcai2022p301} & 98.71 & 80.97 & 42.51 & 95.21 & 79.44 & 68.10\\
 & FedALA~\cite{Zhang_Hua_Wang_Song_Xue_Ma_Guan_2023} & 99.11 & {84.50} & 52.44 & {95.81} & {82.84} & 74.12\\
& pFedHN~\cite{pmlr-v139-shamsian21a} & 98.96 & 84.47 & {52.50} & 95.75 & 82.76 & 74.50\\
& pFedLA~\cite{Ma_2022_CVPR} &98.26 &71.22 &31.22 &92.65 &77.31 &56.09\\[1pt]
\cdashline{2-8}
 & & & & & & & \\[-7pt]
& FedPAC~\cite{xu2023personalized}&97.83&{85.27}&51.40&95.65&81.69&71.79\\ 
  & DBE~\cite{NEURIPS2023_2e0d3c6a} & 99.04 & 84.26 & 51.43 & 95.51 &{82.81} &72.24\\
  & FedProto~\cite{Tan_Long_LIU_Zhou_Lu_Jiang_Zhang_2022} &  {99.14} & 83.28 & 50.16 & 95.45 & 81.73 & {74.86}\\
  & FedGH~\cite{10.1145/3581783.3611781} &  99.07 & 83.37 & 49.84 & 94.87 & 79.61 & 73.04\\
 \cline{2-8}
  & & & & & & & \\[-7pt]
 & \text{\textbf{FedAPA \textit{(ours)}}} & \textbf{99.15} & \textbf{85.33} & \textbf{52.62} & \textbf{95.95} & \textbf{84.14} & \textbf{75.09}\\

\hline 
\end{tabular}
\caption{Test accuracy (\%) of FedAPA and 11 competitors on Fashion MNIST, CIFAR-10, and CIFAR-100 in pathological and practical non-IID settings for 20 and 50 collaborative clients. \textit{c} denotes the number of classes held by each client in pathological settings. $\alpha$ is the concentration parameter of the Dirichlet distribution used in practical settings.}
\label{tab1}
\end{center}
\end{table*}

\section{Experiments}
\subsection{Experimental Setup}
\noindent\textbf{Datasets.} We employ three public image-classification datasets of varying complexities: FMNIST (simple), CIFAR-10 (medium), and CIFAR-100 (complex). Experiments are conducted with 20 and 50 clients, respectively. For each dataset, we construct heterogeneous scenarios following the two widely-used settings: the pathological and the practical settings\cite{pmlr-v54-mcmahan17a,Li_2021_CVPR}. For the pathological setting, we allocate disjoint and unbalanced data of 2/2/10 classes to each client out of a total of 10/10/100 classes on FMNIST/CIFAR-10/CIFAR-100 datasets. For the practical setting, we use a Dirichlet distribution\cite{pmlr-v97-yurochkin19a}, denoted as\textit{ Dir$\left(\alpha\right)$}, to sample data points from each class in a given dataset to each client, where a smaller value of the concentration parameter $\alpha$ reflects stronger data heterogeneity. We set $\alpha$ to 0.1/0.1/0.01 for FMNIST/CIFAR-10/CIFAR-100 datasets.


\noindent\textbf{Competitors.} We evaluated FedAPA against FedAvg, the standard non-PFL baseline, and 10 state-of-the-art PFL algorithms. The comparison includes six personalized model-based  methods (FedPHP, FedAMP, APPLE, FedALA, pFedHN), two global model-based approaches (FedPAC and DBE), and two other PFL approaches (FedProto, and FedGH). The dashed lines in Table \ref{tab1} and Table \ref{tab2} sequentially divide the competitor methods into these three groups.

\noindent\textbf{Model and Training.}
In all experiments, each client uses the same LeNet-5 architecture, with the first four layers serving as shared feature extraction layers with the server, and the final fully connected (FC) layer acting as a private decision layer. To accommodate the size and color channels of the images in each dataset, we set the input channels/FC layer input dimension/output channels of each client model to 1/256/10 for FMNIST, 3/400/10 for CIFAR-10, and 3/400/100 for CIFAR-100. We utilize the default training-to-test set ratio of each dataset, namely 6:1 for FMNIST and 5:1 for CIFAR-10 and CIFAR-100. 
Unless otherwise specified, we use the following training settings. we use SGD with a momentum of 0.9 as the client optimizer. We run two local training epochs in each iteration with a batch size of 64 and a learning rate $\alpha$ = 0.01 for 50 communication rounds with the server. Each  round involves participation from at least 60\% of all users. For our FedAPA algorithm, we initialize the self-weight $\mu$ in aggregation-weight vectors to 0.5 and set the learning rate $\eta$ for updating weight vectors to 0.01.

\noindent\textbf{Implementation.}
The experiments were conducted on a workstation running Ubuntu 22.04 LTS. The implementation utilized Python 3.8.10, CUDA version 11.3, and the PyTorch 1.11.0 deep learning framework. The workstation features a 12 vCPU Intel Xeon Platinum 8255C @ 2.50GHz CPU and an NVIDIA GeForce RTX 2080 Ti GPU.

\subsection{Accuracy Evaluation}
To evaluate and compare the performance of FedAPA with the competitors, we ran each FL algorithm and computed the average accuracy from three random data allocations. Accuracy was calculated as the ratio of correct to total predictions. Table 1 shows the test accuracy of all methods across 12 scenarios, covering three datasets, two client numbers, and two non-IID settings. It also includes the number of classes \textit{c} per client for the pathological settings, and the concentration parameter \textit{$\alpha$} of the Dirichlet distribution for the practical settings. 
As shown in Table \ref{tab1}, FedAPA outperforms all competitors in test accuracy across 10 out of 12 scenarios. The only exceptions are the pathological non-IID settings on the FMNIST and CIFAR-100 datasets with 20 clients, where FedAPA trails the top-performing methods by just 0.06\% (vs. FedPAC) and 0.51\% (vs. FedALA), respectively. FedAPA demonstrates a more pronounced performance advantage in practical non-IID settings, which better reflect realistic heterogeneous data distributions. Overall, FedAPA consistently exhibits superior performance compared to all competing methods.


\subsection{Computation and Communication Overhead}
The computation times for each FL algorithm are summarized in the computation column of Table \ref{tab2}, showing total time (in minutes) and time per iteration (in seconds) for 20 clients using the medium CIFAR-10 dataset in the practical non-IID setting with 2 training epochs. As shown, FedAPA requires 17.66 (approximately 18) seconds per iteration, comparable to the baseline FedAvg. Thus, FedAPA incurs only an additional 0.34 seconds per iteration while achieving significant accuracy improvements. It offers state-of-the-art performance with minimal computation overhead among all compared PFL algorithms.

The communication costs per iteration are listed in the communication column of Table \ref{tab2}. Prototype-based methods, such as FedProto and FedGH, incur the lowest communication overhead by transmitting prototypes instead of model parameters. FedAPA, while achieving higher accuracy, maintains communication efficiency by transmitting only partial model parameters, resulting in the least communication overhead among all non-prototype FL algorithms. In summary, FedAPA achieves superior accuracy with the lowest communication cost among non-prototype methods.



\begin{table}[t]
\begin{center}
\fontsize{7.5pt}{8pt}\selectfont
\begin{tabular}{l|c|c c|c }
\hline
\multirow{4}{*}{\textbf{Approach}}&\multirow{4}{*}{\textbf{Acc.}}&\multicolumn{2}{c|}{\textbf{Comput.}}&\textbf{Commun.}\\
\cline{3-5}
 & & \multirow{2}{*}{Total} & \multirow{2}{*} {Time}  &\multirow{2}{*} {(Practice)}\\[2pt]
 &\textit{} &time &per iter.  &per iter. \\
&\textit{\%} &\textit{min.}  &\textit{sec.}  &\textit{KB} \\
\hline
 & & & & \\[-7pt]
FedAvg~\cite{pmlr-v54-mcmahan17a}  & 56.68  & \textbf{ \textit{42}} & \textbf{\textit{17}}  &484\\
FedPHP~\cite{10.1007/978-3-030-86486-6_36}  &82.52 & 50 & 20 &484\\
FedAMP~\cite{Huang_Chu_Zhou_Wang_Liu_Pei_Zhang_2021}  &82.38 & 46& 19   &484\\
APPLE~\cite{ijcai2022p301}   &79.79 & 50 & 20  &5087\\
FedALA\cite{Zhang_Hua_Wang_Song_Xue_Ma_Guan_2023}   &82.53 &67 & 27 &484\\
pFedHN~\cite{pmlr-v139-shamsian21a}  &80.71&48& 19 &478\\
pFedLA~\cite{Ma_2022_CVPR} &82.86&89&35&484\\
\hdashline 
 & & & &  \\[-7pt]
FedPAC~\cite{xu2023personalized}&84.93 &161 &64 &553\\
DBE~\cite{NEURIPS2023_2e0d3c6a}   & 83.45& 44& 18 &484\\
FedProto~\cite{Tan_Long_LIU_Zhou_Lu_Jiang_Zhang_2022}   &82.71 & 66& 26 &\textbf{5}\\
FedGH~\cite{10.1145/3581783.3611781}  & 80.75 &63& 25 &\textbf{5}\\
\hline
 & & & &\\[-7pt]
\textbf{FedAPA \textit{(ours)}} & \textbf{85.85} & \textbf{44}&\textbf{18}  &\textbf{478}\\
\hline
\end{tabular}
\caption{Comparison of test accuracy (\%), computation time (minutes/seconds), and communication overhead (parameters per iteration per client) on CIFAR-10 (20 clients, practical Non-IID setting, $\alpha = 0.1$).}
\label{tab2}
\end{center}
\end{table}

\subsection{Ablation Study}
(1) FedAPA incorporates the \textbf{Adaptive Personalized Aggregation (APA)} strategy and the \textbf{Partial Model Sharing (PMS)} strategy. To assess the contributions of these two strategies, we conduct an ablation study, as shown in the top half of Table \ref{tab:ablation}. Starting with FedAvg as the baseline, which uses average aggregation and entire-model sharing, we achieve an accuracy of 56.68\%. Introducing the APA strategy significantly improves accuracy by 28.25\%, reaching 84.93\%. Adding the PMS strategy further enhances accuracy by 0.92\%, resulting in a final accuracy of 85.85\%. These results underscore the pivotal contribution of the APA strategy as the primary driver of performance improvement, while the PMS strategy offers a smaller, supplementary boost. 
(2) Following the server-side gradient update of the weight vectors, FedAPA applies three post-processing steps to the weights: clipping, self-weight adjustment, and normalization. Ablation studies in the lower half of Table \ref{tab:ablation} show that removing each step individually reduces accuracy by 2.58\%, 1.44\%, and 5.51\%, respectively, with a significant 26.14\% drop when all are omitted. These results highlight the importance of each step in improving model accuracy.

\begin{table}[t]
\centering

\fontsize{8.4}{10}\selectfont  
\begin{tabular}{l l r}
\toprule
\textbf{Method} & \textbf{$\Delta$Acc.(\%)} & \textbf{Acc.(\%)} \\ 
\midrule
Baseline \textit{ (FedAvg)} & \textbf{--} & 56.68 \\ 
Baseline \textbf{+ APA} & $\uparrow$ \textbf{28.25} & 84.93 \\
Baseline + APA \textbf{+ PMS} \textit{ (FedAPA)} 
& $\uparrow$ 28.25 \textbf{+ 0.92} & 85.85 \\ 
\midrule
\midrule
FedAPA \textit{(ours)}  & \textbf{--} & 85.85 \\ 
w/o \textbf{Clipping} & $\downarrow$ \textbf{2.58} & 83.27 \\
w/o \textbf{Self-weight}  & $\downarrow$ \textbf{1.44} & 84.41 \\ 
w/o \textbf{Normalization} & $\downarrow$ \textbf{5.51} & 80.34 \\
w/o \textbf{All} & $\downarrow$ \textbf{26.14} & 59.71 \\
\bottomrule
\end{tabular}
\caption{Ablation results of 2 strategies and 3 post-processing steps to FedAPA's accuracy (Acc.) on CIFAR-10 (20 clients, practical Non-IID setting).} 
\label{tab:ablation}
\end{table}


\begin{figure}[t]
    \centering
    \includegraphics[width=0.8\linewidth]{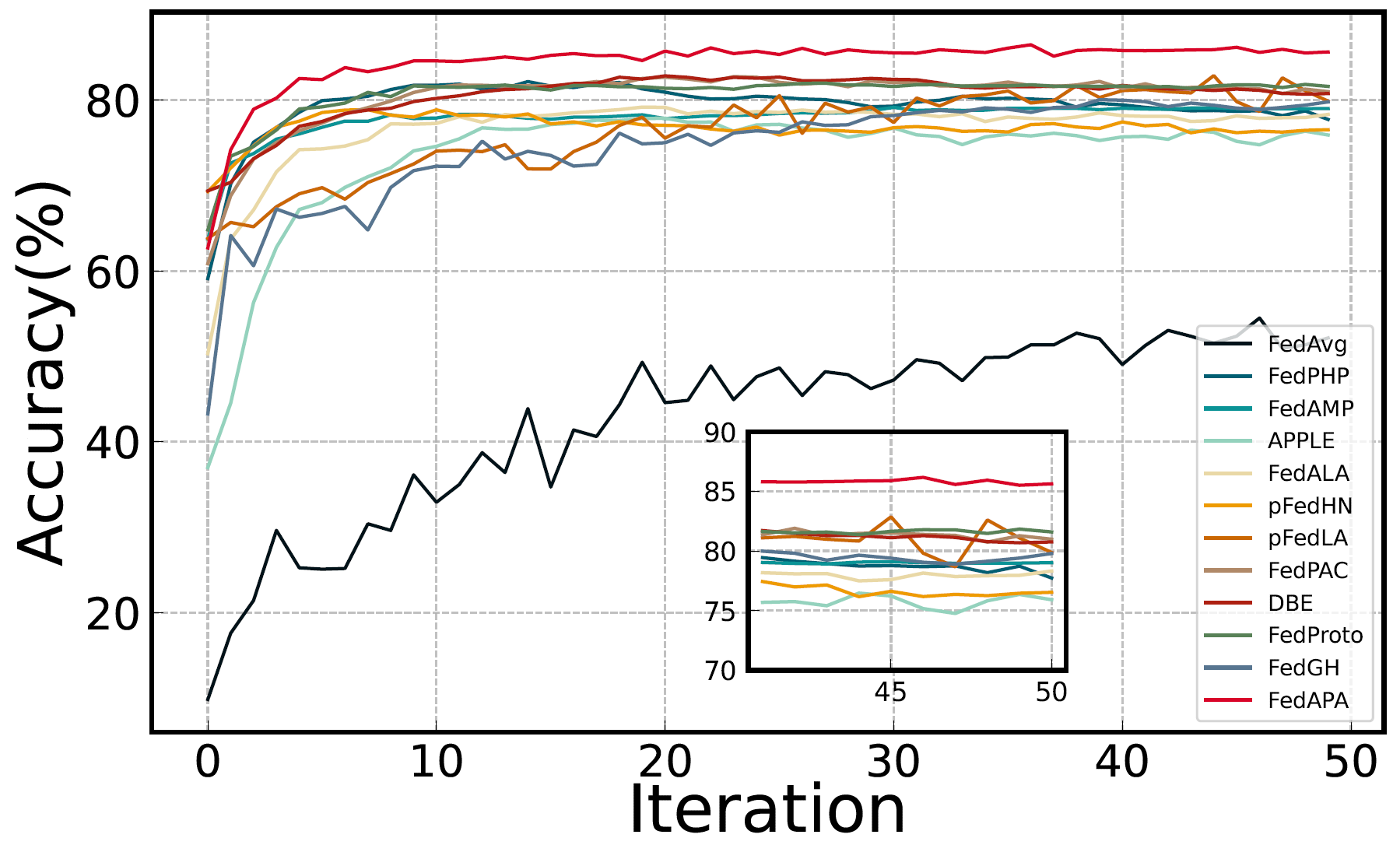}
    \caption{Convergence curves of FedAPA and 11 compared methods on CIFAR-10 (20 clients, practical Non-IID setting).}
    \captionsetup{justification=centering}
    \label{ConvergeFig.png}
\end{figure}
\subsection{Convergence Curve Analysis}
The convergence curves in Fig\ref{ConvergeFig.png} show the performance of the 12 methods over training rounds. FedAPA reaches convergence within approximately 10 rounds, demonstrating its rapid convergence rate compared to the competing methods. The curve for FedAPA exhibits a smooth progress towards convergence, indicating stable training and consistent performance improvement. Upon convergence, FedAPA achieves an accuracy of around 86\%, which is higher than all other competitors. These results confirm that FedAPA is one of the fastest-converging algorithms, achieving the highest accuracy with stable convergence behavior.

\begin{table}[t]
\begin{center}

\fontsize{8.4}{10}\selectfont  

\begin{tabular}{l|c c c c c}
\hline
$\mu$ & 0  &0.3 & \textbf{0.5} & 0.7 & 1\\
\hline
Accuracy & 84.41 & 84.76 & \textbf{85.85} & 84.02 & 82.95\\
\hline
drop &1.44  & 1.09 & \textbf{-} & 1.83 & 2.90\\
\hline
\end{tabular}
\caption{Accuracy of FedAPA with varying self-weight $\mu$ on CIFAR-10 (20 clients, practical Non-IID setting).}
\label{tab3}
\end{center}
\end{table}

\subsection{Effect of Self-weight $\mu$}
FedAPA adjusts the self-weight of each aggregation-weight vector using the hyperparameter $\mu$. We evaluate the effect of $\mu$ on accuracy by varying it from 0 to 1, using 20 clients on CIFAR-10 in a Practical Non-IID setting. As shown in Table \ref{tab3}, the best accuracy is achieved when $\mu = 0.5$. Lower values underweight local knowledge, while higher values overly emphasize local tasks, reducing the benefits of shared knowledge. Thus, $\mu = 0.5$ balances local and shared contributions effectively.

\section{Conclusion}
In this paper, we introduce a new PFL framework, FedAPA, designed to automatically consolidate desired knowledge from collaborative clients to form personalized local models. FedAPA adaptively learns aggregation weights on the server by utilizing gradients derived from changes in client parameters. We provide a theoretical proof for the convergence of our FedAPA algorithm. Experimental results under heterogeneous data scenarios demonstrate that FedAPA achieves higher accuracy and lower computational overhead compared to the competing PFL methods, while maintaining communication efficiency. These contributions highlight the effectiveness and efficiency of FedAPA in improving PFL performance, suggesting its potential suitability for handling large-scale heterogeneous data in practical applications.
\nocite{guo2020novel, guo2023graph, lu2023decomposed, guo2020dual, guo2021conservative, zhou2021device, ijcai2022p311, guo2019adaptive, zhou2021octo, guo2016improved, guo2024multimodal, zhang2023towards, li2024tsca, liu2022towards, guo2019ams, zhou2022cadm, guo2019ee, lu2023drpt, guo2023mdenet, guo2021learning, ma2019position, rao2024srcd, rao2023srcd, wang2023data, wang2024towards, huo2023offline, wang2022exploring, huo2022procc, huo2024utdnet, huo2023utdnet, liu20232, guo2022application, wang2022efficient, guo2022fed, qi2023hwamei, li2023dissecting, zhou2024pass, qi2023arena, wang2023towards, huo2024non, liu2023gbe, guo2023cns, wang2023sfp, guo2024fine, huo2024c2kd, huo2024procc, ijcai2024p449, li2024freepih, rao2023attribute, tang2023learning, zhou2024robustness, li2025personalized, li2024personalized, chen2024fine, wang2024sfp, chen2024vision, ijcai2024p534, bai2024diprompt, chen2025unveiling, chen2024review, chen2024cgraphnet, liu2024epsilon, 
guo2024element, tang2024causally}

\bibliographystyle{named}
\bibliography{ijcai25}

\begin{thebibliography}{}

\bibitem[\protect\citeauthoryear{Ha \bgroup \em et al.\egroup
  }{2016}]{DBLP:journals/corr/HaDL16}
David Ha, Andrew~M. Dai, and Quoc~V. Le.
\newblock Hypernetworks.
\newblock {\em CoRR}, abs/1609.09106, 2016.

\bibitem[\protect\citeauthoryear{Huang \bgroup \em et al.\egroup
  }{2021}]{Huang_Chu_Zhou_Wang_Liu_Pei_Zhang_2021}
Yutao Huang, Lingyang Chu, Zirui Zhou, Lanjun Wang, Jiangchuan Liu, Jian Pei,
  and Yong Zhang.
\newblock Personalized cross-silo federated learning on non-iid data.
\newblock {\em Proceedings of the AAAI Conference on Artificial Intelligence},
  35(9):7865--7873, May 2021.

\bibitem[\protect\citeauthoryear{Li \bgroup \em et al.\egroup
  }{2021a}]{Li_2021_CVPR}
Qinbin Li, Bingsheng He, and Dawn Song.
\newblock Model-contrastive federated learning.
\newblock In {\em Proceedings of the IEEE/CVF Conference on Computer Vision and
  Pattern Recognition (CVPR)}, pages 10713--10722, June 2021.

\bibitem[\protect\citeauthoryear{Li \bgroup \em et al.\egroup
  }{2021b}]{10.1007/978-3-030-86486-6_36}
Xin-Chun Li, De-Chuan Zhan, Yunfeng Shao, Bingshuai Li, and Shaoming Song.
\newblock Fedphp: Federated personalization with inherited private models.
\newblock In Nuria Oliver, Fernando P{\'e}rez-Cruz, Stefan Kramer, Jesse Read,
  and Jose~A. Lozano, editors, {\em Proceedings of the Joint European
  Conference on Machine Learning and Knowledge Discovery in Databases. Research
  Track}, pages 587--602, Cham, 2021. Springer International Publishing.

\bibitem[\protect\citeauthoryear{Luo and Wu}{2022}]{ijcai2022p301}
Jun Luo and Shandong Wu.
\newblock Adapt to adaptation: Learning personalization for cross-silo
  federated learning.
\newblock In Lud~De Raedt, editor, {\em Proceedings of the Thirty-First
  International Joint Conference on Artificial Intelligence, {IJCAI-22}}, pages
  2166--2173. International Joint Conferences on Artificial Intelligence
  Organization, 7 2022.
\newblock Main Track.

\bibitem[\protect\citeauthoryear{Ma \bgroup \em et al.\egroup
  }{2022}]{Ma_2022_CVPR}
Xiaosong Ma, Jie Zhang, Song Guo, and Wenchao Xu.
\newblock Layer-wised model aggregation for personalized federated learning.
\newblock In {\em Proceedings of the IEEE/CVF Conference on Computer Vision and
  Pattern Recognition (CVPR)}, pages 10092--10101, June 2022.

\bibitem[\protect\citeauthoryear{McMahan \bgroup \em et al.\egroup
  }{2017}]{pmlr-v54-mcmahan17a}
Brendan McMahan, Eider Moore, Daniel Ramage, Seth Hampson, and Blaise Aguera~y
  Arcas.
\newblock {Communication-Efficient Learning of Deep Networks from Decentralized
  Data}.
\newblock In Aarti Singh and Jerry Zhu, editors, {\em Proceedings of the 20th
  International Conference on Artificial Intelligence and Statistics},
  volume~54 of {\em Proceedings of Machine Learning Research}, pages
  1273--1282. PMLR, 20--22 Apr 2017.

\bibitem[\protect\citeauthoryear{Shamsian \bgroup \em et al.\egroup
  }{2021}]{pmlr-v139-shamsian21a}
Aviv Shamsian, Aviv Navon, Ethan Fetaya, and Gal Chechik.
\newblock Personalized federated learning using hypernetworks.
\newblock In Marina Meila and Tong Zhang, editors, {\em Proceedings of the 38th
  International Conference on Machine Learning}, volume 139 of {\em Proceedings
  of Machine Learning Research}, pages 9489--9502. PMLR, 18--24 Jul 2021.

\bibitem[\protect\citeauthoryear{Tan \bgroup \em et al.\egroup
  }{2022}]{Tan_Long_LIU_Zhou_Lu_Jiang_Zhang_2022}
Yue Tan, Guodong Long, LU~LIU, Tianyi Zhou, Qinghua Lu, Jing Jiang, and Chengqi
  Zhang.
\newblock Fedproto: Federated prototype learning across heterogeneous clients.
\newblock {\em Proceedings of the AAAI Conference on Artificial Intelligence},
  36(8):8432--8440, Jun. 2022.

\bibitem[\protect\citeauthoryear{Wu \bgroup \em et al.\egroup }{2020}]{9090366}
Qiong Wu, Kaiwen He, and Xu~Chen.
\newblock Personalized federated learning for intelligent iot applications: A
  cloud-edge based framework.
\newblock {\em IEEE Open Journal of the Computer Society}, 1:35--44, 2020.

\bibitem[\protect\citeauthoryear{Xu \bgroup \em et al.\egroup
  }{2023}]{xu2023personalized}
Jian Xu, Xinyi Tong, and Shao-Lun Huang.
\newblock Personalized federated learning with feature alignment and classifier
  collaboration.
\newblock {\em arXiv preprint arXiv:2306.11867}, 2023.

\bibitem[\protect\citeauthoryear{Yang \bgroup \em et al.\egroup
  }{2019}]{10.1145/3298981}
Qiang Yang, Yang Liu, Tianjian Chen, and Yongxin Tong.
\newblock Federated machine learning: Concept and applications.
\newblock {\em ACM Trans. Intell. Syst. Technol.}, 10(2), jan 2019.

\bibitem[\protect\citeauthoryear{Yi \bgroup \em et al.\egroup
  }{2023}]{10.1145/3581783.3611781}
Liping Yi, Gang Wang, Xiaoguang Liu, Zhuan Shi, and Han Yu.
\newblock Fedgh: Heterogeneous federated learning with generalized global
  header.
\newblock In {\em Proceedings of the 31st ACM International Conference on
  Multimedia}, MM '23, page 8686–8696, New York, NY, USA, 2023. Association
  for Computing Machinery.

\bibitem[\protect\citeauthoryear{Yurochkin \bgroup \em et al.\egroup
  }{2019}]{pmlr-v97-yurochkin19a}
Mikhail Yurochkin, Mayank Agarwal, Soumya Ghosh, Kristjan Greenewald, Nghia
  Hoang, and Yasaman Khazaeni.
\newblock {B}ayesian nonparametric federated learning of neural networks.
\newblock In Kamalika Chaudhuri and Ruslan Salakhutdinov, editors, {\em
  Proceedings of the 36th International Conference on Machine Learning},
  volume~97 of {\em Proceedings of Machine Learning Research}, pages
  7252--7261. PMLR, 09--15 Jun 2019.

\bibitem[\protect\citeauthoryear{Zhang \bgroup \em et al.\egroup
  }{2023}]{Zhang_Hua_Wang_Song_Xue_Ma_Guan_2023}
Jianqing Zhang, Yang Hua, Hao Wang, Tao Song, Zhengui Xue, Ruhui Ma, and
  Haibing Guan.
\newblock Fedala: Adaptive local aggregation for personalized federated
  learning.
\newblock {\em Proceedings of the AAAI Conference on Artificial Intelligence},
  37(9):11237--11244, Jun. 2023.

\bibitem[\protect\citeauthoryear{Zhang \bgroup \em et al.\egroup
  }{2024}]{NEURIPS2023_2e0d3c6a}
Jianqing Zhang, Yang Hua, Jian Cao, Hao Wang, Tao Song, Zhengui XUE, Ruhui Ma,
  and Haibing Guan.
\newblock Eliminating domain bias for federated learning in representation
  space.
\newblock In A.~Oh, T.~Naumann, A.~Globerson, K.~Saenko, M.~Hardt, and
  S.~Levine, editors, {\em Advances in Neural Information Processing Systems},
  volume~36, pages 14204--14227. Curran Associates, Inc., 2024.
\newblock [Online]. Available:
  \url{https://dl.acm.org/doi/10.5555/3666122.3666747}.

\end{thebibliography}

\newpage
\Large{\textbf{APPENDIX}}
\appendix
\section{CONVERGENCE ANALYSIS}

\noindent\textbf{Additional Notation}

Here, additional notations are introduced to better represent the process of local model update. Let $f_i(\theta_{i})$ denote the embedding function for the $i$-th client, which may vary across different clients. The decision function for all clients is $h_i(z_i)$. Thus, the labeling function can be written as $F_i(\theta_{i}, \phi_{i}) = h_i(\phi_i) \circ f_i(\theta_{i})$, and sometimes we use $\omega_i$ to represent $[\theta_i; \phi_i]$ for brevity. Let $A_i$ denote the aggregated weight vector of user $i$, and let $a_{i, j}$ denote the $j$-$th$ learnable weight parameter of user $i$'s aggregation weight, where $A_i$ = $[$$a_{i, 1}$, $a_{i, 2}$, $\dots$, $a_{i, \textit{M}}$$]^T$, for $i,j$ = 1, 2, $\dots$, \textit{M}. Therefore, the local loss function of client $i$ can be written as:
\begin{equation}
\begin{aligned}
    \mathcal{L}(\omega_{i}; x, y) = \mathcal{L}(h(\phi_i; f(\theta_{i}; x)), y)
    \label{con:1}
\end{aligned}
\end{equation}
we use $t$ to represent the communication round and $e\in\{\frac{1}{2},1,2,\dots,E\}$ to represent the local iterations. There are $E$ local iterations in total, so $tE + e$ refers to the $e$-th local iteration in the communication round $t + 1$. Moreover, $tE$ represents the time step before the aggregation of client parameters, and $tE + 1/2$ represents the time step between parameters aggregation and the first iteration of the current round.

\noindent \textbf{Key Lemmas}

\noindent \textbf{Lemma 1.}\hypertarget{L:1}{}
Let Assumption \hyperlink{A:1}{1} and Assumption \hyperlink{A:1}{2} in the main text hold. From the beginning of communication round $t + 1$ to the last local update step, the loss function of an arbitrary client can be bounded as:
\begin{equation}
\begin{aligned}
    \mathbb{E}[\mathcal{L}_{(t + 1)E}] \le 
    &\mathbb{E}[\mathcal{L}_{tE + \frac{1}{2}}] + \frac{L_1 E \alpha ^2}{2} \sigma^2\\
    &- (\alpha - \frac{L_1 \alpha ^2}{2}) \sum_{e = \frac{1}{2}}^{E - 1} || \nabla \mathcal{L}_{tE + e} ||_2^2
\end{aligned}
\end{equation}
Proof. Due to the fact that this lemma applies to an arbitrary client, the client notation $i$ is omitted. Let $\omega ^{(t+1)} = \omega^{(t)} - \alpha gr^{(t)}$, then
\begin{equation}
\begin{aligned}
     \mathcal{L}_{tE + 1}  
     &\overset{(a)}{\le} \mathcal{L}_{tE + \frac{1}{2}} + \langle \nabla \mathcal{L}_{tE + \frac{1}{2}},(\omega^{(tE + 1)} - \omega^{(tE + \frac{1}{2})}) \rangle  \\
     &\hspace{5mm} + \frac{L_1}{2}||\omega^{(tE + 1)} - \omega^{(tE + \frac{1}{2})}||_2^2\\
     &= \mathcal{L}_{tE + \frac{1}{2}} - \alpha \langle \nabla \mathcal{L}_{tE + \frac{1}{2}}, gr^{(tE + \frac{1}{2})} \rangle\\
     &\hspace{5mm} + \frac{L_1}{2}|| \alpha gr^{(tE + \frac{1}{2})} ||_2^2,
\end{aligned}
\end{equation}
where $(a)$ follows from the quadratic $L_1$-$Lipschitz$ smooth bound in Assumption \hyperlink{A:1}{1} in the main text. Taking expectation of both sides of the above equation on the random variable $\xi^{(tE+\frac{1}{2})}$, we have
\begin{equation}
\begin{aligned}
     \mathbb{E}[\mathcal{L}_{tE + 1}]  
     &\le \mathbb{E}[\mathcal{L}_{tE + \frac{1}{2}}] - \alpha \mathbb{E}[ \langle \nabla \mathcal{L}_{(tE + \frac{1}{2})}, gr^{(tE + \frac{1}{2})} \rangle]  \\
     &\hspace{5mm}+ \frac{L_1 \alpha ^2}{2} \mathbb{E}[|| gr^{(tE + \frac{1}{2})} ||_2^2]\\
     &\overset{(b)}{=} \mathbb{E}[\mathcal{L}_{tE + \frac{1}{2}}] - \alpha || \nabla \mathcal{L}_{tE + \frac{1}{2}} ||_2^2  \\
     &\hspace{5mm}+ \frac{L_1 \alpha ^2}{2} \mathbb{E}[|| gr^{(tE + \frac{1}{2})} ||_2^2]\\
     &\overset{(c)}{\le} \mathbb{E}[\mathcal{L}_{tE + \frac{1}{2}}] - \alpha || \nabla \mathcal{L}_{tE + \frac{1}{2}} ||_2^2  \\
     &\hspace{5mm} + \frac{L_1 \alpha ^2}{2} (|| \nabla \mathcal{L}_{tE + \frac{1}{2}} ||_2^2 + Var(gr^{(tE + \frac{1}{2})}))\\
     &= \mathbb{E}[\mathcal{L}_{tE + \frac{1}{2}}] - (\alpha - \frac{L_1 \alpha ^2}{2})|| \nabla \mathcal{L}_{tE + \frac{1}{2}} ||_2^2 \\
     &\hspace{5mm}+ \frac{L_1 \alpha ^2}{2} Var(gr^{(tE + \frac{1}{2})}) \\
     &\overset{(d)}{\le} \mathbb{E}[\mathcal{L}_{tE + \frac{1}{2}}] - (\alpha - \frac{L_1 \alpha ^2}{2})|| \nabla \mathcal{L}_{tE + \frac{1}{2}} ||_2^2 \\
     &\hspace{5mm}+ \frac{L_1 \alpha ^2}{2} \sigma^2
\end{aligned}
\end{equation}
where $(b)$ and $(d)$ follow from Assumption \hyperlink{A:1}{2} in the main text, and $(c)$ follows from $Var(x) = \mathbb{E}[x^2] - (\mathbb{E}[x])^2$. Then, by telescoping over $E$ steps, we have
\begin{equation}
\begin{aligned}
    \mathbb{E}[\mathcal{L}_{(t + 1)E}] 
    &\le \mathbb{E}[\mathcal{L}_{tE + \frac{1}{2}}] + \frac{L_1 E \alpha ^2}{2} \sigma^2\\
    & - (\alpha - \frac{L_1 \alpha ^2}{2}) \sum_{e = \frac{1}{2}}^{E - 1} || \nabla \mathcal{L}_{tE + e} ||_2^2
\end{aligned}
\end{equation}

\noindent \textbf{Lemma 2.}\hypertarget{L:2}{}
For any $t = 1, 2, \dots$,we have
\begin{equation}
\begin{aligned}
    ||(A_i^{(t + 1)} - e_i)|| \le ||(A_i^{(0)} - e_i)|| + 2\eta Max\sum_{k = 0}^t|| (\Theta^{(k)})^T||
\end{aligned}
\end{equation}
Proof. (The norm $||\cdot||$ appearing in this lemma refers to the $L_2$-norm $||\cdot||_2$.)
\begin{equation}
\begin{aligned}
    ||&(A_i^{(t + 1)} - e_i)|| \\
    &\overset{(a)}{=} ||A_i^{(t)} - \eta (\nabla_{A_i} \bar{\theta}_{i}^{(t + 1)})^T(\theta_{i}^{(t + 1)} - \bar{\theta_{i}}^{(t + 1)}) - e_i||\\
    &= ||(A_i^{(t)} - e_i) - \eta (\nabla_{A_i} \bar{\theta}_{i}^{(t + 1)})^T(\theta_{i}^{(t + 1)} - \bar{\theta_{i}}^{(t + 1)})||\\
    &\overset{(b)}{\le} ||(A_i^{(t)} - e_i)|| + ||\eta (\nabla_{A_i} \bar{\theta}_{i}^{(t + 1)})^T(\theta_{i}^{(t + 1)} - \bar{\theta_{i}}^{(t + 1)})|| \\
    &\overset{(c)}{\le} ||(A_i^{(t)} - e_i)|| + 2\eta Max|| (\nabla_{A_i} \bar{\theta}_{i}^{(t + 1)})^T|| \\
    &\overset{(d)}{\le} ||(A_i^{(t)} - e_i)|| + 2\eta Max|| (\Theta^{(t)})^T|| \\
    \label{con:14}
\end{aligned}
\end{equation}
Since inequality $(\ref{con:14})$ holds, for any $t = 1, 2, \dots$,
\begin{equation}
\begin{aligned}
    ||(A_i^{(t + 1)} - e_i)|| \le ||(A_i^{(0)} - e_i)|| + 2\eta Max\sum_{k = 0}^t|| (\Theta^{(k)})^T||
\end{aligned}
\end{equation} 
where $Max = \underset{t = 1, 2, \dots}{\underset{i = 1, 2, \dots, M}{max}} ||\theta_i^{(t)}||$, $\Theta^{(t)} = (\theta_1^{(t)}, \theta_2^{(t)}, \dots, \theta_M^{(t)})$, $e_i = (0, \dots, 1, \dots, 0)^{T}$ with 1 at the $i$-$th$ position. Here, $(a)$ follows from Eq. $(\ref{con:2})$ and Eq. $(\ref{Formula (7)})$, $(b)$ and $(c)$ follow from $||a - b|| \le ||a|| + ||b||$, and $(d)$ follows from $\bar{\theta}_{i}^{(t + 1)} = \Theta^{(t)}A_i^{(t)}$.

\noindent \textbf{Lemma 3.}\hypertarget{L:3}{}
Let Assumption \hyperlink{A:1}{3} in the main text hold. After the parameter aggregation at the server, the loss function of an arbitrary client can be bounded as:
\begin{equation}
\begin{aligned}
    \mathbb{E}[\mathcal{L}_{(t + 1)E + \frac{1}{2}}] \le \mathbb{E}[\mathcal{L}_{(t + 1)E}] + 2L_2\eta(t + 1) Max^3
\end{aligned}
\end{equation}
Proof. (The norm $||\cdot||$ appearing in this lemma refers to the $L_2$-norm $||\cdot||_2$.)
\begin{equation}
\begin{aligned}
        \mathcal{L}_{(t + 1)E + \frac{1}{2}}
        &= \mathcal{L}_{(t + 1)E} + \mathcal{L}_{(t + 1)E + \frac{1}{2}} - \mathcal{L}_{(t + 1)E} \\
        &\overset{(a)}{\le} \mathcal{L}_{(t + 1)E} + L_2||\theta_{i}^{((t + 1)E + \frac{1}{2})} - \theta_{i}^{((t + 1)E)}|| \\
        &=  \mathcal{L}_{(t + 1)E} + L_2||\bar{\theta}^{(t + 2)}_i - \theta^{(t + 1)}_i|| \\
        &\overset{(b)}{=} \mathcal{L}_{(t + 1)E} + L_2||\sum_{j = 1}^M a_{i, j}\theta_{j}^{(t + 1)} - \theta_i^{(t + 1)}|| \\
        &= \mathcal{L}_{(t + 1)E} + L_2||\Theta^{(t + 1)}A_i^{(t + 1)} - \theta_i^{(t + 1)}|| \\
        &\overset{(c)}{\le} \mathcal{L}_{(t + 1)E} + L_2||\Theta^{(t + 1)}(A_i^{(t + 1)} - e_i)|| \\
        &\overset{(d)}{\le} \mathcal{L}_{(t + 1)E} + L_2Max||(A_i^{(t + 1)} - e_i)|| \\
        &\overset{(e)}{\le} \mathcal{L}_{(t + 1)E} + L_2Max(||(A_i^{(0)} - e_i)||  \\
        &\hspace{19mm}+ 2\eta Max\sum_{k = 0}^t|| (\Theta^{(k)})^T||)\\
        &\overset{(f)}{\le} \mathcal{L}_{(t + 1)E} + 2L_2\eta (t + 1)Max^3 \\
\end{aligned}
\end{equation}
Taking expectations of random variable $\xi$ on both sides, then
\begin{equation}
\begin{aligned}
        \mathbb{E}[\mathcal{L}_{(t + 1)E + \frac{1}{2}}]
         &\le \mathbb{E}[\mathcal{L}_{(t + 1)E}] + 2L_2\eta (t + 1)Max^3
\end{aligned}
\end{equation}  
where $Max = \underset{t = 1, 2, \dots}{\underset{i = 1, 2, \dots, M}{max}} ||\theta_i^{(t)}||$,  $\Theta^{(t)} = (\theta_1^{(t)}, \theta_2^{(t)}, \dots, \theta_M^{(t)})$, $e_i = (0, \dots, 1, \dots, 0)^{T}$ with 1 at the $i$-th position. Here, $(a)$  follows from $L2$-$Lipschitz$ continuity in Assumption \hyperlink{A:1}{3} in the main text, $(b)$ follows from Eq. $(\ref{con:2})$ in the main text, $(c)$ follows from $\theta_i^{(t + 1)} = \Theta^{(t + 1)}e_i$, $(d)$ follows from $||\Theta^{(t + 1)}|| \le Max $, $(e)$ follows from lemma \hyperlink{L:2}{2}, and $(f)$ follows from $A_i^{(0)} = e_i$.

\noindent \textbf{Theorem}

\noindent\textbf{Theorem 1.} Let Assumption 1 to 3 in the main text hold. For an arbitrary client, after every communication round, we have
\begin{equation}
\begin{aligned}
        \mathbb{E}[\mathcal{L}_{(t + 1)E + \frac{1}{2}}] 
        &\le \mathbb{E}[\mathcal{L}_{tE + \frac{1}{2}}] - (\alpha - \frac{L_1 \alpha ^2}{2}) \sum_{e = \frac{1}{2}}^{E - 1} || \nabla \mathcal{L}_{tE + e} ||_2^2 \\
        &+ \frac{L_1 E \alpha ^2}{2} \sigma^2 + 2L_2\eta (t + 1)Max^3
\end{aligned}
\end{equation}  
where $Max = \underset{t = 1, 2, \dots}{\underset{i = 1, 2, \dots, M}{max}} ||\theta_i^{(t)}||$.

\noindent Proof. Combining Lemma \hyperlink{L:1}{1} and lemma \hyperlink{L:3}{3}, we easily obtain
\begin{equation}
\begin{aligned}
        \mathbb{E}[\mathcal{L}_{(t + 1)E + \frac{1}{2}}] 
        &\le \mathbb{E}[\mathcal{L}_{tE + \frac{1}{2}}] - (\alpha - \frac{L_1 \alpha ^2}{2}) \sum_{e = \frac{1}{2}}^{E - 1} || \nabla \mathcal{L}_{tE + e} ||_2^2 \\
        &+ \frac{L_1 E \alpha ^2}{2} \sigma^2 + 2L_2\eta (t + 1)Max^3
\end{aligned}
\end{equation}



\noindent \textbf{Corollary 1.}
The loss function $\mathcal{L}$ for any arbitrary client exhibits a monotonic decrease with each communication round when
\begin{equation}
\begin{aligned}
        \alpha
        &< 4L_2\eta (t + 1)Max^3\Big(\sum_{e = \frac{1}{2}}^{E - 1} || \nabla \mathcal{L}_{tE + e} ||_2^2\\
        &\hspace{3mm}- ((\sum_{e = \frac{1}{2}}^{E - 1} || \nabla \mathcal{L}_{tE + e} ||_2^2)^2 \\
        &\hspace{3mm}- 4L_1L_2 \eta (t+1)Max^3(\sum_{e = \frac{1}{2}}^{E - 1} || \nabla \mathcal{L}_{tE + e} ||_2^2 + E\sigma^2))^\frac{1}{2}\Big)^{-1}
\end{aligned}
\end{equation}  
and
\begin{equation}
\begin{aligned}
        \alpha
        &> 4L_2\eta (t + 1)Max^3\Big(\sum_{e = \frac{1}{2}}^{E - 1} || \nabla \mathcal{L}_{tE + e} ||_2^2\\
        &\hspace{3mm}+ ((\sum_{e = \frac{1}{2}}^{E - 1} || \nabla \mathcal{L}_{tE + e} ||_2^2)^2 \\
        &\hspace{3mm}- 4L_1L_2 \eta (t+1)Max^3(\sum_{e = \frac{1}{2}}^{E - 1} || \nabla \mathcal{L}_{tE + e} ||_2^2 + E\sigma^2))^\frac{1}{2}\Big)^{-1}
\end{aligned}
\end{equation}  
and
\begin{equation}
\begin{aligned}
        \eta
        &< \frac{(\sum_{e = \frac{1}{2}}^{E - 1} || \nabla \mathcal{L}_{tE + e} ||_2^2)^2}{4L_1L_2(t + 1)Max^3(\sum_{e = \frac{1}{2}}^{E - 1} || \nabla \mathcal{L}_{tE + e} ||_2^2 + E\sigma^2)}
\end{aligned}
\end{equation}  

Hence, the convergence of $\mathcal{L}$ is proven.

\end{document}